\pdfoutput=1
\documentclass[10pt,twocolumn,letterpaper]{article}

\usepackage{iccv_copyright}
\usepackage{times}
\usepackage{epsfig}
\usepackage{graphicx}
\usepackage{amsmath}
\usepackage{amssymb}
\usepackage{algorithm}
\usepackage{algpseudocode}
\usepackage{multirow}
\usepackage{booktabs}
\usepackage{subfigure}
\usepackage[percent]{overpic}

\usepackage[pagebackref=true,breaklinks=true,letterpaper=true,colorlinks,bookmarks=false]{hyperref}

\newcommand{\argmin}[1]{\underset{#1}{\operatorname{argmin}}}
\def\lenfunc{\operatorname{len}}

\def\SSS{$\mathcal{S}3$\xspace}

\def\III{$\mathcal{I}3$\xspace}
\def\II{$\mathcal{I}2$\xspace}
\def\I{$\mathcal{I}1$\xspace}
\def\modelS{$\mathcal{S}$\xspace}
\def\modelI{$\mathcal{I}$\xspace}

\def\OurAlgo{Intrinsic Architecture Search\xspace}

\def\OurResNet{ResiaNet\xspace}
\def\OurDetNet{DetiaNet\xspace}
\def\OurResNetSSS{ResiaNet\,\SSS}
\def\OurResNetIII{ResiaNet\,\III}
\def\OurResNetI{ResiaNet\,\I}

\def\OurDetNetII{DetiaNet\,\II}

\def\sMACs{{\small{ (MACs)}}\xspace}
\def\sparams{{\small{ (params)}}\xspace}

\iccvfinalcopy %
\ieeecopyright
\def\ieeeCopyrightYear{2019}
\def\ieeeCopyrightTextPre{ICCV 2019 Workshop on Neural Architects (Oral).}

\def\iccvPaperID{****} %

\begin{document}

\title{Understanding the Effects of Pre-Training for Object Detectors\\via Eigenspectrum}

\author{Yosuke Shinya\\
DENSO CORPORATION\\
{\tt\small yosuke.shinya.j7r@jp.denso.com}
\and
Edgar Simo-Serra\\
Waseda University\\
{\tt\small ess@waseda.jp}
\and
Taiji Suzuki\\
The University of Tokyo / RIKEN\\
{\tt\small taiji@mist.i.u-tokyo.ac.jp}
}

\maketitle

\begin{abstract}
   ImageNet pre-training has been regarded as essential for training accurate object detectors for a long time. Recently, it has been shown that object detectors trained from randomly initialized weights can be on par with those fine-tuned from ImageNet pre-trained models. However, the effects of pre-training and the differences caused by pre-training are still not fully understood. In this paper, we analyze the eigenspectrum dynamics of the covariance matrix of each feature map in object detectors. Based on our analysis on ResNet-50, Faster R-CNN with FPN, and Mask R-CNN, we show that object detectors trained from ImageNet pre-trained models and those trained from scratch behave differently from each other even if both object detectors have similar accuracy. Furthermore, we propose a method for automatically determining the widths (the numbers of channels) of object detectors based on the eigenspectrum. We train Faster R-CNN with FPN from randomly initialized weights, and show that our method can reduce $\sim$27\% of the parameters of ResNet-50 without increasing Multiply-Accumulate operations and losing accuracy. Our results indicate that we should develop more appropriate methods for transferring knowledge from image classification to object detection (or other tasks).
\end{abstract}

\section{Introduction}

Object detection and instance segmentation are important tasks, with many real-world applications in robotics, healthcare, etc.
Up until now, most detection and segmentation tasks have relied on ImageNet~\cite{ImageNet_IJCV2015} fine-tuning~\cite{R-CNN, Yosinski2014HowTA}.
With fine-tuning, learned parameters or features of source tasks may be forgotten after learning target tasks~\cite{kirkpatrick2017overcoming},
and domain similarity between tasks being important for transfer learning~\cite{Taskonomy}.
Furthermore, transferring knowledge between dissimilar tasks may cause negative transfer~\cite{NegativeTransfer_rosenstein2005, Wang2019CharacterizingAA}.
Thus many works have discussed
task difference between image classification and object detection~\cite{YOLO9000, DSOD, SNIP_Singh_2018_CVPR, RevisitingRCNN_Cheng_2018_ECCV}
and the effects of pre-training for object detectors~\cite{Huh2016WhatMI, Sun2017RevisitingUE, Mahajan_2018_ECCV, Li2019AnAO}. However, the influence caused by the task difference is still an open problem,
and what and how to transfer knowledge from image classification to object detection are unclear.
Avoiding these problems, it was recently shown that models trained on COCO~\cite{MicrosoftCOCO} from random initialization can be on par with models trained (fine-tuned) from ImageNet pre-trained models~\cite{He2018Scratch}, but it is not clear whether or not models with similar performance have different properties.
To further understand the effects of fine-tuning object detectors, we analyze the eigenspectrum dynamics of the covariance matrix of each feature map in object detectors, and propose a method to automatically determine the numbers of channels necessary for performance. (Each feature map includes channel dimension in this paper.)

\begin{figure}[t]
	\begin{center}
		\begin{overpic}[width=1.0\linewidth]{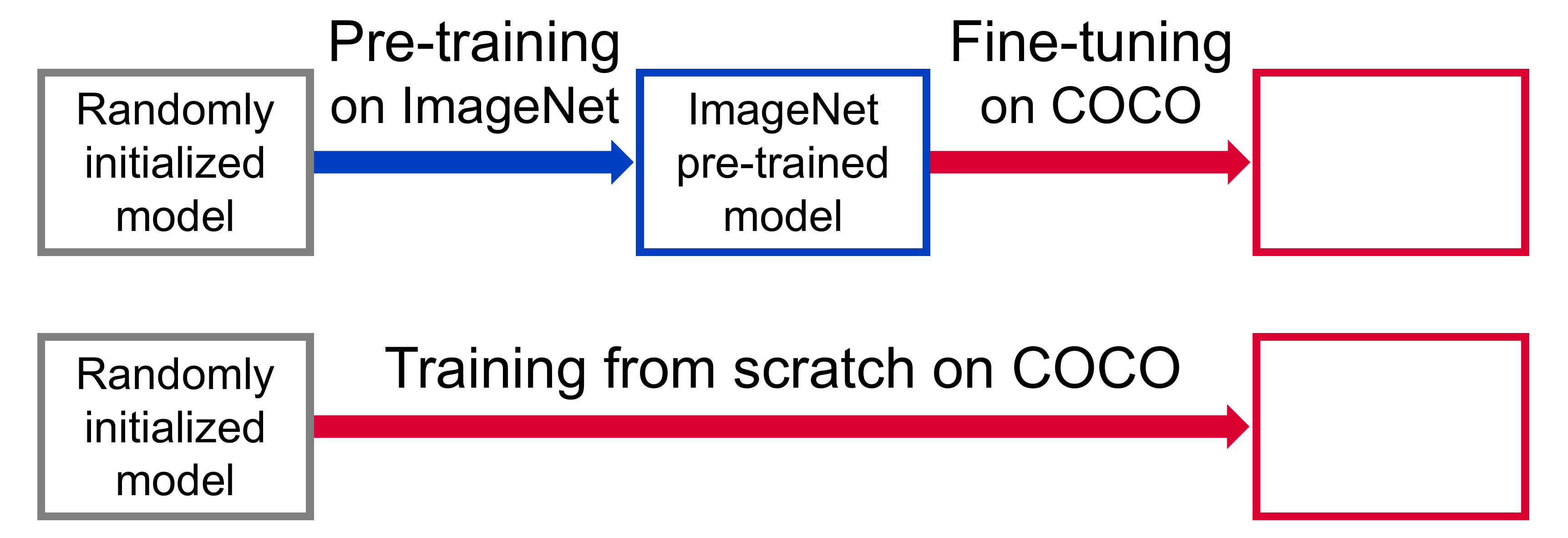}
			\put(86.8,21.7){\Large{\modelI}}
			\put(86.6, 4.9){\Large{\modelS}}
		\end{overpic}
	\textsf{\small{
		Do object detectors \modelI and \modelS converge to similar models?
	}}
	\end{center}
	\caption{
		R-CNN~\cite{R-CNN} shows that
		object detectors \modelI, which fine-tuned from ImageNet pre-trained models, can achieve high accuracy.
		``Rethinking ImageNet Pre-training''~\cite{He2018Scratch} shows that
		object detectors \modelS, which trained from scratch, can achieve similar accuracy to \modelI under appropriate conditions.
		In this paper,
		we show that
		\modelI and \modelS behave differently from each other even if both object detectors have similar accuracy.
	}
	\label{fig:finetuning_vs_scratch}
\end{figure}

More specifically, motivated by the accurate object detectors trained from scratch~\cite{He2018Scratch, ScratchDet},
we focus on the following research question.
\textit{
Do object detectors fine-tuned from ImageNet pre-trained models and those trained from scratch converge to similar models?
}
If the answer is ``Yes,''
we will have a better understanding of the task difference and the behavior of deep neural networks,
and if the answer is ``No, these object detectors do not converge to similar models, but show similar accuracy by chance,'' 
we should incorporate the benefits of both object detectors.

To answer this question,
we train object detectors as shown in Figure~\ref{fig:finetuning_vs_scratch},
and analyze the redundancy of feature maps in the detectors.
To be more precise, we analyze the intrinsic dimensionalities of the feature maps,
which represent how much information the feature maps memorize.
Intrinsic dimensionalities can be quantified by
calculating the eigenspectra of the covariance matrices of the feature maps,
and are related to generalization error~\cite{Suzuki2018FastGE}.
In this paper, we use the numbers of eigenvalues greater than a threshold
as a simple metric of intrinsic dimensionalities,
and we call the sets of intrinsic dimensionalities in a certain network the \textit{intrinsic architecture}.

Our contributions are as follows.
\vspace{-0.4\baselineskip}
\begin{itemize}
	\setlength{\itemsep}{-0.05mm}
	\item
	We analyze the eigenspectrum dynamics of the covariance matrix of each feature map in object detectors,
	and show that
	object detectors trained from ImageNet pre-trained models and those trained from scratch
	behave differently from each other even if both object detectors have similar accuracy.

	\item We propose a method for automatically determining the widths (the numbers of channels) of object detectors.
	We report the results of Faster R-CNN with FPN trained from scratch,
	and show that our method can reduce $\sim$27\% of the parameters of ResNet-50
	without increasing Multiply-Accumulate operations (MACs) and losing accuracy,
	and can improve COCO AP by 0.3\% without increasing parameters (See Sec.~\ref{sec:efficiency_on_coco}).

	\item We explain why architectures and learning schedules of prior object detectors trained from scratch work well in Sec.~\ref{sec:understanding_prior_work}.
	We bridge recent theoretical analysis on generalization error~\cite{Suzuki2018FastGE}
	and the experimental results of recent object detectors~\cite{He2018Scratch}.
\end{itemize}

\section{Related Work}

\subsection{Neural Network Generalization}

One of the most important mysteries of neural networks is its generalization ability.
To understand it,
some work has discussed the relation between generalization and compressibility~\cite{ShwartzZiv2017OpeningTB, PWCCA, Arora2018StrongerGB, Suzuki2018FastGE}.
Information Bottleneck~\cite{ShwartzZiv2017OpeningTB, michael2018on} and Canonical Correlation Analysis (CCA)~\cite{SVCCA, PWCCA} are used for analyzing the dynamics of neural networks.
\cite{ShwartzZiv2017OpeningTB} showed that training with Stochastic Gradient Descent (SGD) has two phases (a label fitting phase and a representation compression phase).
\cite{michael2018on} shows that
networks with ReLU do not necessarily exhibit the compression phase, and that
fitting to task-relevant information and the compression of task-irrelevant information occur simultaneously.
\cite{PWCCA} shows that
generalizing/larger networks converge to more similar solutions than memorizing/smaller networks.
Using CCA, Transfusion~\cite{Transfusion} analyzes the effects of pre-training for classifying medical images,
which are clearly different from natural images in ImageNet and COCO.

The most related theoretical analysis to this paper is the
\textit{degree of freedom} of reproducing kernel Hilbert spaces (RKHSs), which is defined in~\cite{Suzuki2018FastGE}.
Suzuki~\cite{Suzuki2018FastGE} shows the following two important properties of neural networks which motivate our work.
(i) ``if the eigenvalues of the kernels decreases rapidly, then
the degree of freedom gets smaller, and we achieve a better generalization by using a simpler model.''
(ii) ``the effective dimension of the network is less than the actual number of parameters.''
Spectral-Pruning~\cite{Suzuki2018SpectralPruningCD}, which uses the degree of freedom~\cite{Suzuki2018FastGE} as the intrinsic dimensionality of models, is applicable to compress complicated networks.
Our method and analysis are based on eigenspectrum~\cite{Suzuki2018FastGE, Suzuki2018SpectralPruningCD}
and the dynamics of neural networks~\cite{ShwartzZiv2017OpeningTB, SVCCA, PWCCA}.
However, these prior works do not analyze the behavior of neural networks when fine-tuned for object detection 
from ImageNet pre-trained models.

\begin{figure*}[t]
	\vspace{-2mm}
	\begin{center}
		\begin{overpic}[width=0.8\linewidth]{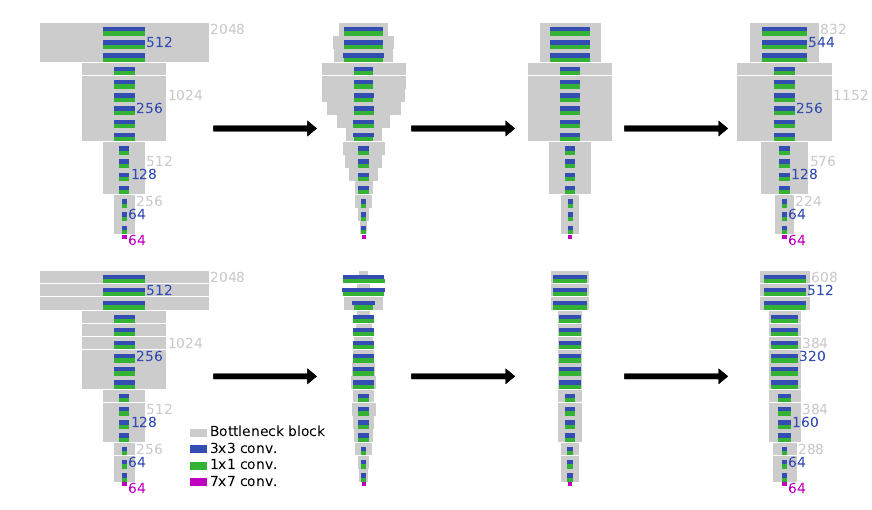}
			\put( 1.7, 54.0){\Large{\modelI}}
			\put( 1.5, 25.5){\Large{\modelS}}
			\put(25.5, 41.0){\small{\textsf{Shrinking}}}
			\put(25.5, 12.6){\small{\textsf{Shrinking}}}
			\put(48.3, 41.0){\small{\textsf{Adjusting}}}
			\put(48.3, 12.6){\small{\textsf{Adjusting}}}
			\put(72.1, 41.0){\small{\textsf{Expanding}}}
			\put(72.1, 12.6){\small{\textsf{Expanding}}}
			\put( 9.5, 30.6){\small{(a)}}
			\put(37.5, 30.6){\small{(b)}}
			\put(60.5, 30.6){\small{(c)}}
			\put(85.0, 30.6){\small{(d)}}
			\put( 9.5,  2.2){\small{(e)}}
			\put(37.5,  2.2){\small{(f)}}
			\put(60.5,  2.2){\small{(g)}}
			\put(85.0,  2.2){\small{(h)}}
		\end{overpic}
	\end{center}
	\vspace{-4mm}
	\caption{
		Overview of \OurAlgo.
		The width of each rectangle is proportional to the output width of each layer.
		7$\times$7, 1$\times$1, and 3$\times$3 convolutions are denoted in purple, green, and blue respectively,
		and bottleneck building blocks are denoted in gray like~\cite{MorphNet}.
		(a): The ResNet-50 backbone of an object detector \modelI fine-tuned from an ImageNet pre-trained model.
		(e): The ResNet-50 backbone of an object detector \modelS trained from scratch.
		(a) and (e) have the same (extrinsic) architecture, and \modelI and \modelS have similar accuracy.
		Thus the difference between \modelI and \modelS is unclear.
		To clarify it, we shrink (by extracting intrinsic architectures), adjust, and expand widths.
		See Sec.~\ref{sec:algorithm} for the details of our algorithm.
		See Sec.~\ref{sec:exp_intrinsic_architecture} for the details of models used in our experiments.
	}
	\label{fig:method}
\end{figure*}
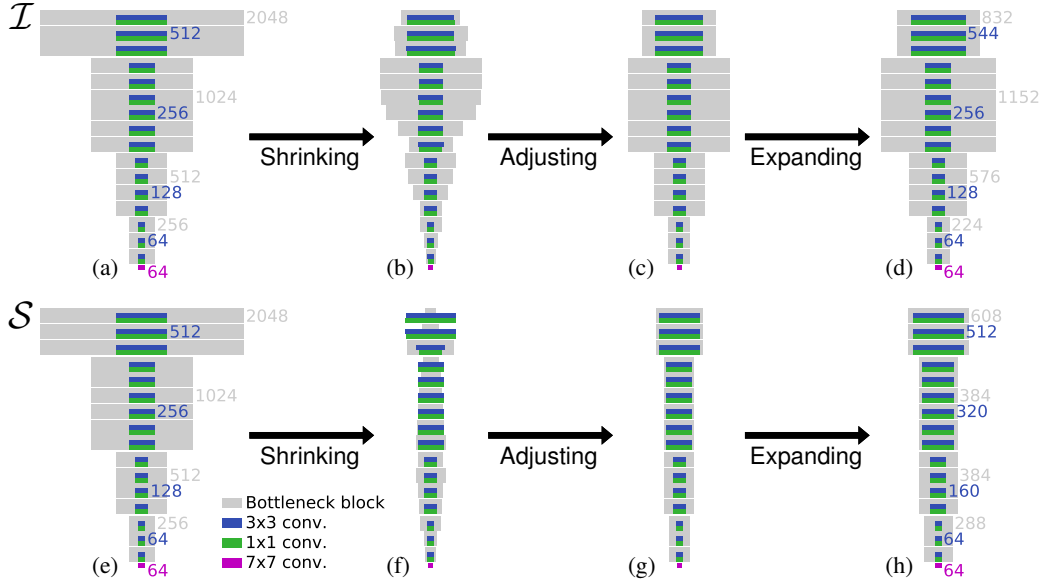

\subsection{Neural Architecture Search (NAS)}

NAS has been a hot research topic on deep learning
since the success of NAS with reinforcement learning~\cite{NAS_Zoph2017},
and efficient methods have broadened its applicability~\cite{NASNet, DARTS, ProxylessNAS}.
Genetic CNN~\cite{GeneticCNN} and NASNet~\cite{NASNet}
transfer architectures learned on a proxy dataset (\eg, CIFAR-10) to large-scale datasets (\eg, ImageNet).
On the other hand, ProxylessNAS~\cite{ProxylessNAS} reduces memory consumption by path-level binarization,
and directly learns architectures for a large-scale dataset.
In addition to NAS for image classification,
a few works focus NAS for semantic segmentation~\cite{ConvolutionalNeuralFabrics, Chen2018SearchingFE, AutoDeepLab} and object detection~\cite{NAS-FPN, DetNAS}.
NAS-FPN~\cite{NAS-FPN} and DetNAS~\cite{DetNAS} search the architectures of Feature Pyramid Networks~\cite{FPN} and backbones for object detectors respectively.
However, these prior works~\cite{NAS-FPN, DetNAS} do not determine the widths of feature maps automatically,
and computational costs for training will become higher if their search space includes the widths.

Determining the widths of feature maps in CNNs can be considered as a subset of NAS.
Although various approaches have been proposed~\cite{Domhan2015_HPO, AMC, ChamNet, MetaPruning},
shrink-and-expand~\cite{MorphNet, NeuralRejuvenation} is a more suitable approach for object detectors because of its simplicity and scalability.
MorphNet~\cite{MorphNet} shrinks and linearly expands networks.
The shrinking imposes L1 regularization on the scaling factors of Batch Normalization to identify and prune unimportant channels
like Network Slimming~\cite{NetworkSlimming_Liu_2017_ICCV},
and takes into account specific resource constraints (\eg, the number of floating point operations).
Neural Rejuvenation~\cite{NeuralRejuvenation} revives dead neurons (reallocates and reinitializes useless channels) during training.
Although the effectiveness of these methods~\cite{MorphNet, NeuralRejuvenation} is verified on ImageNet,
it is unknown whether these methods can be applied to object detectors.

\subsection{Object Detection and Instance Segmentation}

Object detection is one of the core technologies in computer vision,
and has advanced rapidly with deep neural networks~\cite{OverFeat, MultiBox, R-CNN, Fast_R-CNN, Faster_R-CNN_NIPS, YOLO, SSD, FPN, SNIP_Singh_2018_CVPR}
(Refer to a survey~\cite{Liu2018DeepLF} for details).
In addition, instance segmentation~\cite{MNC_Dai_2016_CVPR, FCIS_Li_2017_CVPR, Mask_R-CNN, PANet},
which is the task of segmenting and classifying individual objects,
is important for further detailed object recognition.
Most methods for these tasks train models from ImageNet pre-trained models for better accuracy.
However, pre-training backbones in object detectors on image classification dataset
causes learning bias and limits architecture design~\cite{DSOD, ScratchDet}.

To avoid the problems of pre-training, training object detectors from scratch (from randomly initialized weights) has been discussed in some literature~\cite{DSOD, Tiny-DSOD, DetNet, CornerNet, ScratchDet, He2018Scratch}.
DSOD~\cite{DSOD} shows that
deep supervision~\cite{DeeplySupervisedNets} is critical for training single-shot object detectors from scratch,
and adopts implicit deep supervision via dense connections~\cite{DenseNet}.
ScratchDet~\cite{ScratchDet} shows that Batch Normalization~\cite{BatchNormalization, Santurkar_NeurIPS2018}
helps training from scratch to converge,
and redesigns the backbone of single-shot object detectors.
\cite{He2018Scratch} shows that
Mask R-CNN trained from scratch with appropriate normalization and longer training (instead of pre-training)
can be on par with those fine-tuned from ImageNet pre-trained models.

The most similar work to ours is DetNet~\cite{DetNet}, which is a specialized backbone for object detection.
DetNet mainly focuses on scales (the receptive fields and the spatial resolutions of feature maps)
to overcome drawbacks of ImageNet pre-trained models designed for image classification.
However, its widths are manually determined.
On the other hand, our method does not aim to determine the spatial resolutions.
Using our method and DetNet complementarily would be beneficial.

\section{\OurAlgo}

In this section, we propose a method for automatically determining the widths (the numbers of channels) of feature maps.
Our method reflects intrinsic architectures by calculating the redundancy of feature maps,
and is applicable to complicated networks, such as Faster R-CNN with FPN and Mask R-CNN.
Figure~\ref{fig:method} shows an overview of our method.
We call our algorithm \textit{\OurAlgo},
and we call architectures discovered by our algorithm \textit{\OurResNet} whose base backbone is ResNet.

\subsection{Determining Widths}

Optimizing the widths of feature maps is formulated as
\begin{equation}
O_{1:M}^* = \argmin{c(O_{1:M})\le \zeta}
\min_\theta \mathcal{L}(\theta),
\end{equation}
where $M$ is the total number of layers,
$O_{1:M}$ are the widths of output feature maps,
$\theta$ is the parameters (weights) in neural networks,
$\mathcal{L}$ is a loss function for training neural networks,
$c$ is a function for calculating resource consumption (\eg, Multiply-Accumulate operations (MACs)),
and $\zeta$ is a specified maximum allowable resource consumption.
This formulation is exactly the same as~\cite{MorphNet},
and most notations in this section and some descriptions in Algorithm~\ref{algo:main} follow MorphNet~\cite{MorphNet} for ease of comparing methods.

Although MorphNet~\cite{MorphNet} and Neural Rejuvenation~\cite{NeuralRejuvenation}
also tackle the determination of widths,
these methods need to change training and intrinsic dimensionalities.
In addition, applying them to object detection and instance segmentation poses some difficulties below.
(i) These methods depend on Batch Normalization~\cite{BatchNormalization}.
Therefore, applying them to networks with other normalization layers~\cite{GroupNormalization, SwitchableNormalization} is not trivial. Furthermore, when we apply them to networks without normalization layers~\cite{FixupInitialization}, we need to add Batch Normalization layers~\cite{NeuralRejuvenation}.
(ii) These methods use additional regularizers.
Since object detection and instance segmentation are multi-task learning including classification and localization,
we might need to balance regularization.
(iii) These methods need to train multiple models~\cite{MorphNet} or tune additional hyperparameters~\cite{NeuralRejuvenation}.
This is a serious problem especially for object detection and instance segmentation
because training for these tasks takes a long time (See model zoos of~\cite{Detectron2018, mmdetection, massa2018mrcnn}).

\subsection{Overview}
\label{sec:algorithm}

We propose a method for determining the widths of object detectors using eigenspectrum~\cite{Suzuki2018FastGE}.
Algorithm~\ref{algo:main} shows the whole process,
where $S_{1:M}$ are the eigenspectra of feature maps,
$d_{1:M}$ are the intrinsic dimensionalities of the feature maps,
$T$ is a threshold for calculating intrinsic dimensionalities (\eg, $10^{-3}$),
$\lenfunc$ is a function for counting numbers which meet the condition,
and $\omega$ is a width multiplier.

The details of Algorithm~\ref{algo:main} are described below.
In \textbf{Step 1}, we set initial weights. Weights in a base backbone (\eg, ResNet-50) are initialized from one of the ImageNet pre-trained models, or randomly initialized. Weights out of the base backbone are randomly initialized.
In \textbf{Step 2}, we train the whole network (\eg, Faster R-CNN with FPN or Mask R-CNN) with the base backbone.
In \textbf{Step 3}, we calculate the eigenspectrum of each feature map in the whole network (See Sec.~\ref{sec:calculating_eigenspectra} for details).
Eigenvalues are normalized with the largest eigenvalue of each feature map.
In \textbf{Step 4}, we shrink the widths of each feature map by extracting an intrinsic architecture (See Sec.~\ref{sec:shrinking_widths} for details).
In \textbf{Step 5}, we adjust the widths mainly for networks with multiple branches (See Sec.~\ref{sec:adjusting_widths} for details).
In \textbf{Step 6}, we expand the widths by linear expanding (See Sec.~\ref{sec:expanding_widths} for details).

\begin{algorithm}[t]
	\caption{\OurAlgo}
	\begin{algorithmic}[1]
		\State Set initial weights.
		\State Train the whole network to find
		\Statex $\theta^* = \argmin{\theta}\,\mathcal{L}(\theta)$.
		\State Calculate eigenspectra $S_{1:M}$.
		\State Calculate intrinsic dimensionalities $d_{1:M}$ by
		\Statex $d_{1:M} = \lenfunc(S_{1:M} > T)$.
		\State Determine new widths $O_{1:M}'$ by adjusting $d_{1:M}$.
		\State Find the largest $\omega$ such that $c(\omega \cdot O_{1:M}') \le  \zeta$.\\
		\Return $\omega\cdot O_{1:M}'$.
	\end{algorithmic}
	\label{algo:main}
\end{algorithm}

\subsection{Calculating Eigenspectra}
\label{sec:calculating_eigenspectra}

When we calculate the eigenspectra of feature maps which have spatial resolutions (\ie almost all feature maps in CNNs),
we normalize the covariance matrices by the resolutions.
Specifically, the (non-centered) covariance matrix $\Sigma$ of a feature map $F$ is calculated as
\begin{equation}
\Sigma = \frac{1}{n} \sum_{i=1}^n \frac{1}{W_i H_i} \sum_{x=1}^{W_i} \sum_{y=1}^{H_i} F_{i, x, y} F_{i, x, y}^\top,
\end{equation}
where $n$ is the number of images (we randomly sample 5,000 images from the training set in our experiments),
$W_i, H_i$ are the spatial width and height of the feature map,
and $F_{i, x, y}$ is a feature vector whose coordinates are $(x, y)$ in the feature map for the $i$-th image.
Not only $F_{i, x, y}$ but also $W_i, H_i$ depend on images feed-forwarded because input image resolutions may change in the case of COCO.

We calculate the eigenspectra of feature maps before or after convolutional layers, fully connected layers, and transposed convolutional layers.
Note that feature maps after $L$-th convolutional layer and feature maps before $(L+1)$-th convolutional layer generally do not match due to normalization layers and activation layers.

\subsection{Shrinking Widths}
\label{sec:shrinking_widths}

We calculate intrinsic dimensionalities from the eigenspectra.
We use the numbers of eigenvalues greater than a predefined threshold as intrinsic dimensionalities.
(Using degree of freedom~\cite{Suzuki2018SpectralPruningCD} may be better, though we do not use pruning and we set random values to the initial weights.)
Although we set $10^{-3}$ to the threshold in our experiment, we may get better accuracy if we tuned the threshold as a hyperparameter.

\subsection{Adjusting Widths}
\label{sec:adjusting_widths}

If the network has multiple branches,
adjusting intrinsic dimensionalities is necessary to determine new widths,
because either the input feature maps or the output feature maps of branches may have to have the same widths.
Especially for ResNet with bottlenecks, 
where the widths of feature maps which pass through shortcuts are set to the maximum intrinsic dimensionalities in the same stage for preserving most information which flows shortcut.
Furthermore, we set the same output widths to the first and the second convolutional layers of all residual blocks in the same stage by calculating the geometric mean of intrinsic dimensionalities.
This setting has some advantages:
(i) The second convolutional layers of residual blocks can be replaced with depthwise convolutional layers like~\cite{MobileNetV2}.
(ii) Using the same widths is efficient considering memory access cost~\cite{ShuffleNetV2}.
(iii) Implementation is easy and thus modifications to the code of ResNet are minimized.

\subsection{Expanding Widths}
\label{sec:expanding_widths}

Our expanding method is basically the same as that of~\cite{MorphNet}.
Specifically, the output width of each layer is multiplied by a uniform width multiplier $\omega$ to fit a target resource consumption.
The optimal $\omega$ can be found by a binary search because $c(\omega \cdot O_{1:M}')$ monotonically increases with $\omega$ in our experiments.
$c(O_{1:M})$ is calculated as
\begin{equation}
c(O_{1:M}) = \sum_{L=1}^{M+1} I_L O_L K_L^2 W_L H_L,
\end{equation}
when targeting MACs, and
\begin{equation}
c(O_{1:M}) = \sum_{L=1}^{M+1} I_L O_L K_L^2,
\end{equation}
when targeting the number of parameters,
where
$I_L, O_L$ are the widths of the input/output feature map,
$K_L$ is the kernel size,
and $W_L, H_L$ are the spatial width and height of the output feature map,
for each layer $L = 1, \ldots, M+1$.
We consider $O_{M+1}$ is a fixed number (\eg, 1,000 for ImageNet classification).
For simplicity, we consider the spatial width and height of kernel size to be the same in each layer,
and omit the resource consumption of biases.

To avoid odd widths~\cite{MorphNet},
we round $\omega\cdot O_{1:M}'$ to hardware-friendly multiples (\eg, multiples of 4, 8, 16, or 32).
This rounding is also useful for networks with Group Normalization layers~\cite{GroupNormalization}.
When resource fittings are too coarse, we may fill the gaps by increasing widths greedily.

\begin{figure}[t]
	\begin{center}
		\includegraphics[height=0.48\linewidth]{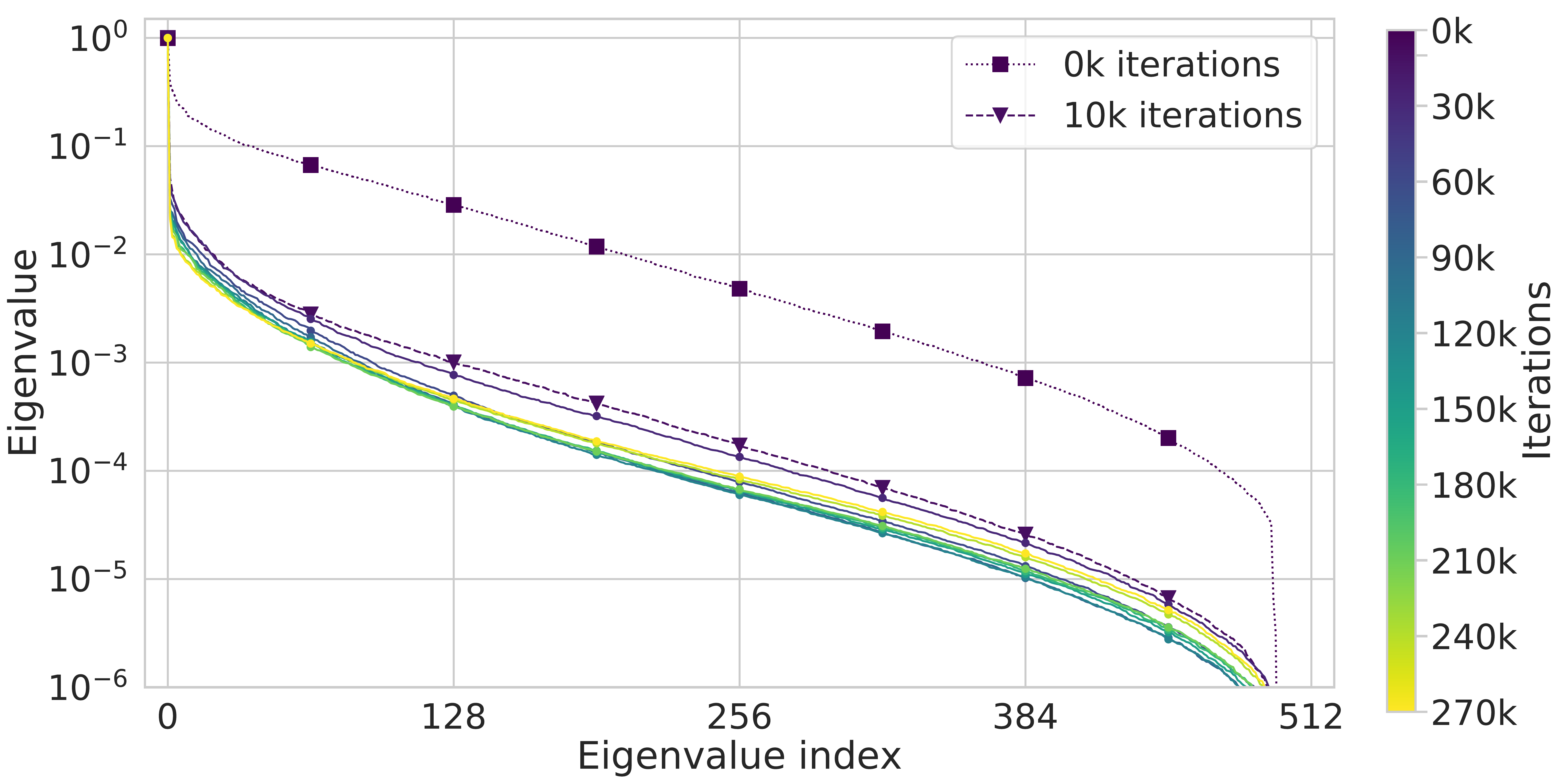}
	\end{center}
	\vspace{-3mm}
	\caption{
		Dropping of eigenvalues.
		Some feature maps in the stage 5 of ResNet-50 become highly redundant in the first 10k iterations,
		though it is irredundant before fine-tuning (0k iterations).
	}
	\label{fig:dynamics_drop}
\end{figure}

\begin{figure}[t]
	\begin{center}
		\includegraphics[height=0.48\linewidth]{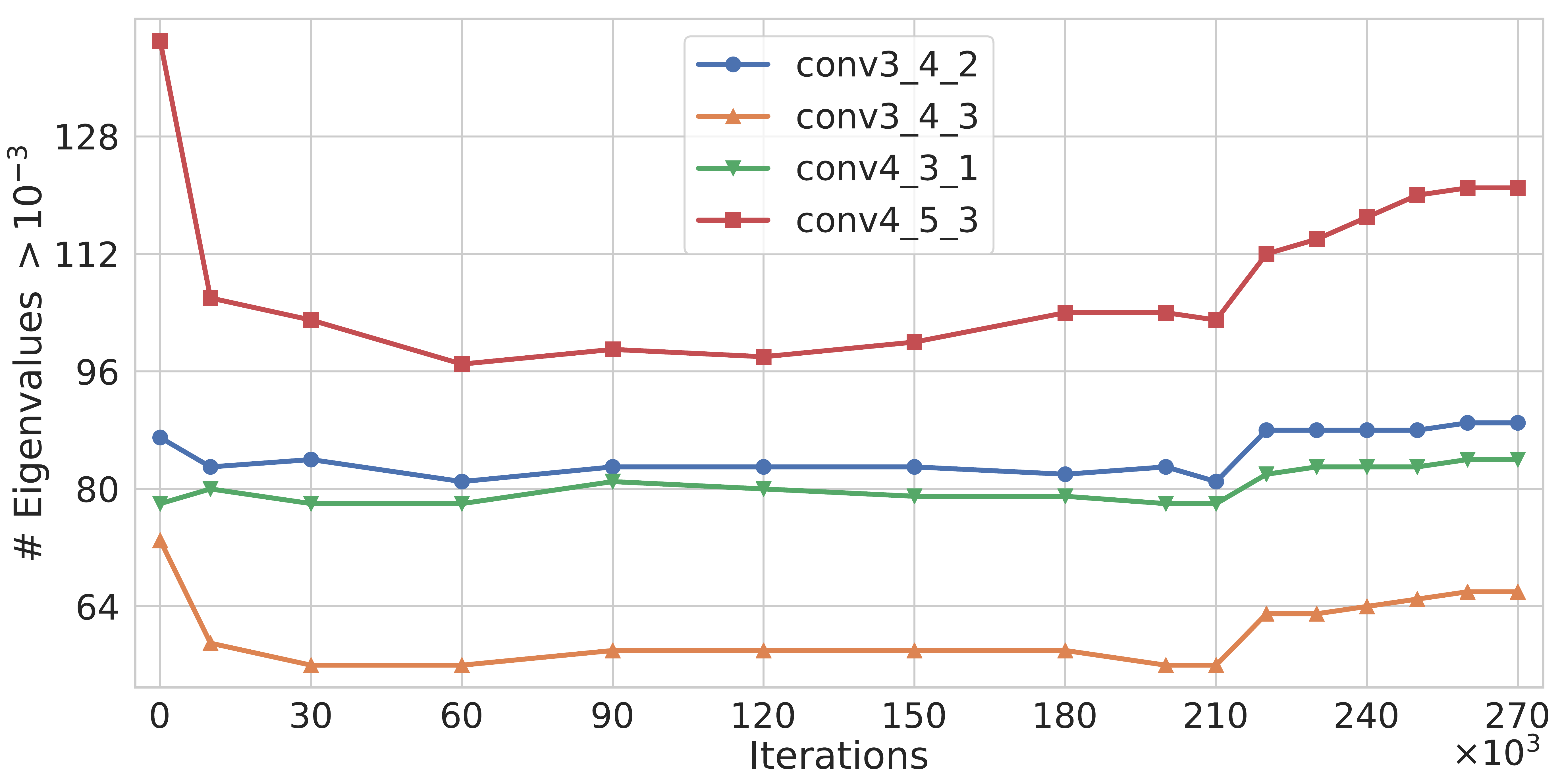}
	\end{center}
    \vspace{-3mm}
	\caption{Rebounding of eigenvalues. The numbers of eigenvalues greater than $10^{-3}$ increase
		immediately after the first learning rate decay (210k iterations) in some feature maps.}
	\label{fig:dynamics_rebound}
\end{figure}

\section{Experiments}
\label{sec:experiments}

To analyze the effects of pre-training for object detectors
and to verify the effectiveness of our method,
we conduct experiments on COCO.

\subsection{Experimental Settings}
\label{sec:expsetting}

The experimental settings mainly follow Mask R-CNN~\cite{Mask_R-CNN}
in Detectron~\cite{Detectron2018} (which includes implementation by the authors of Mask R-CNN)
like~\cite{He2018Scratch}.
Our implementation is based on Detectron.pytorch~\cite{Detectron_pytorch},
which is a PyTorch implementation of Detectron.

We use ResNet-50~\cite{ResNet_CVPR2016} as a base backbone.
We train Faster R-CNN~\cite{Faster_R-CNN_NIPS} with Feature Pyramid Network (FPN)~\cite{FPN}
and Mask R-CNN~\cite{Mask_R-CNN} in an end-to-end manner~\cite{Faster_R-CNN_NIPS}.
We use Group Normalization (GN)~\cite{GroupNormalization},
because appropriate normalization is a key factor for training from scratch~\cite{ScratchDet, He2018Scratch},
and GN has several advantages~\cite{GroupNormalization} compared to Synchronized Batch Normalization~\cite{MegDet}.
The learning rate settings follow~\cite{He2018Scratch}. Specifically, 
the initial learning rate is 0.02 with warm-up~\cite{Goyal2017AccurateLM},
and the learning rate is reduced by $10\times$.
Iterations for the first decay, the second decay, and ending training are
60k, 80k, 90k for $1\times$ schedule,
120k, 160k, 180k for $2\times$ schedule, and
210k, 250k, 270k for $3\times$ schedule.
We use synchronous SGD
with an effective batch size of 16 (= 2 images/GPU $\times$ 8 GPUs),
a momentum of 0.9, and a weight decay of $10^{-4}$.

All models are trained on COCO \texttt{train2017} set (118,287 images)
and evaluated on COCO \texttt{val2017} set (5,000 images) with COCO metrics unless otherwise stated.

\begin{figure*}[t]
	\begin{center}
		\includegraphics[height=0.24\linewidth]{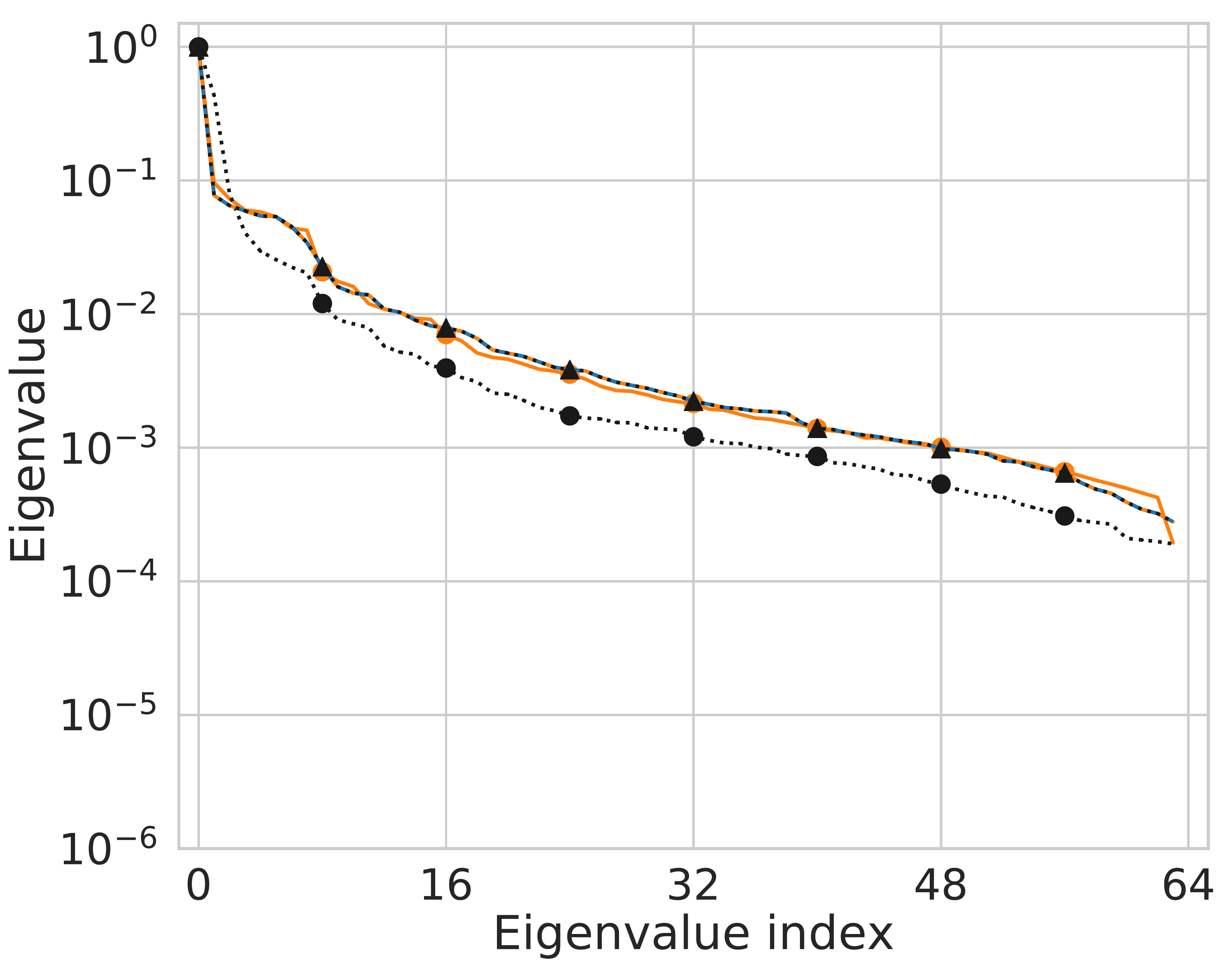}
		\hspace{2mm}
		\includegraphics[height=0.24\linewidth]{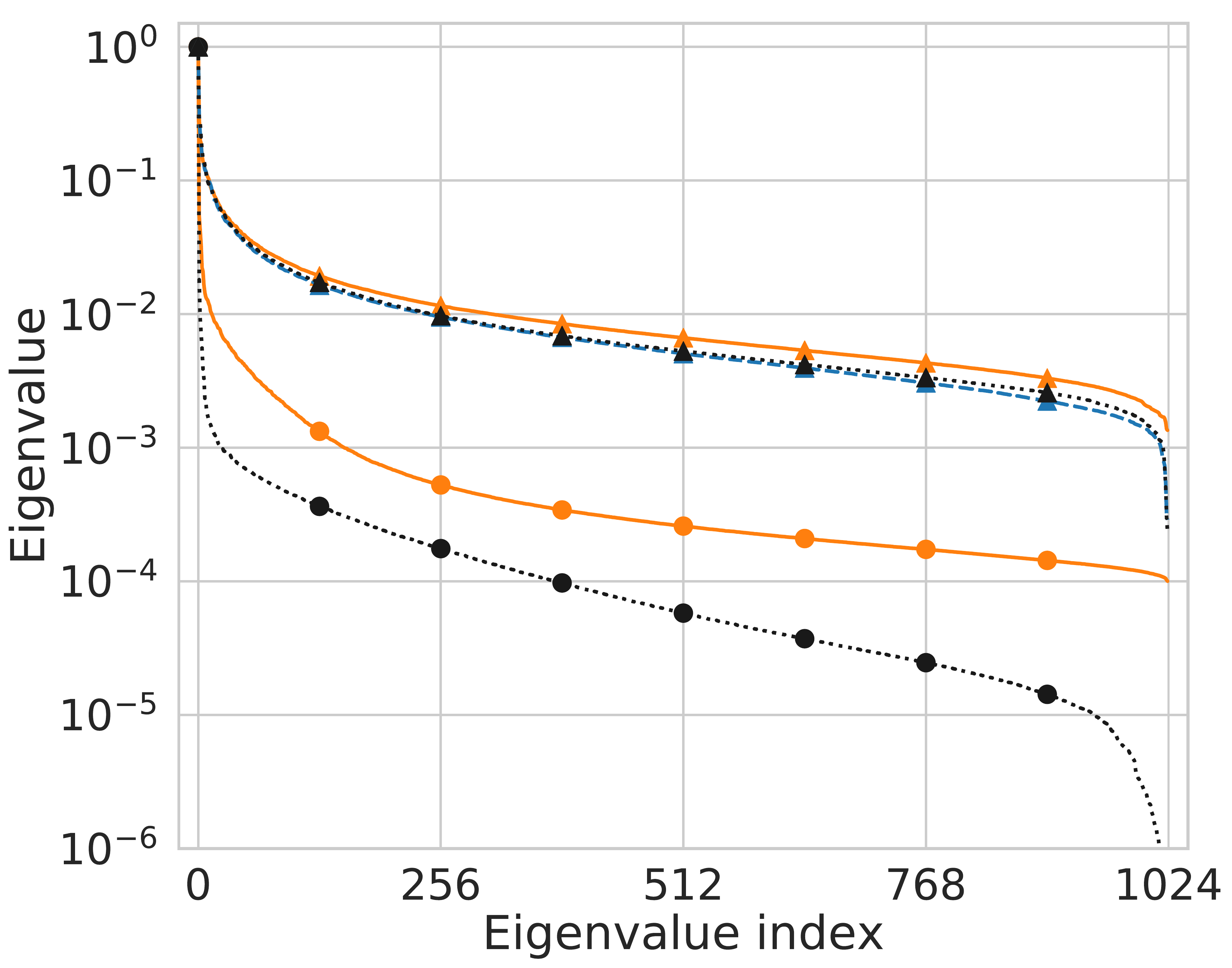}
		\hspace{1.1mm}
		\includegraphics[height=0.24\linewidth]{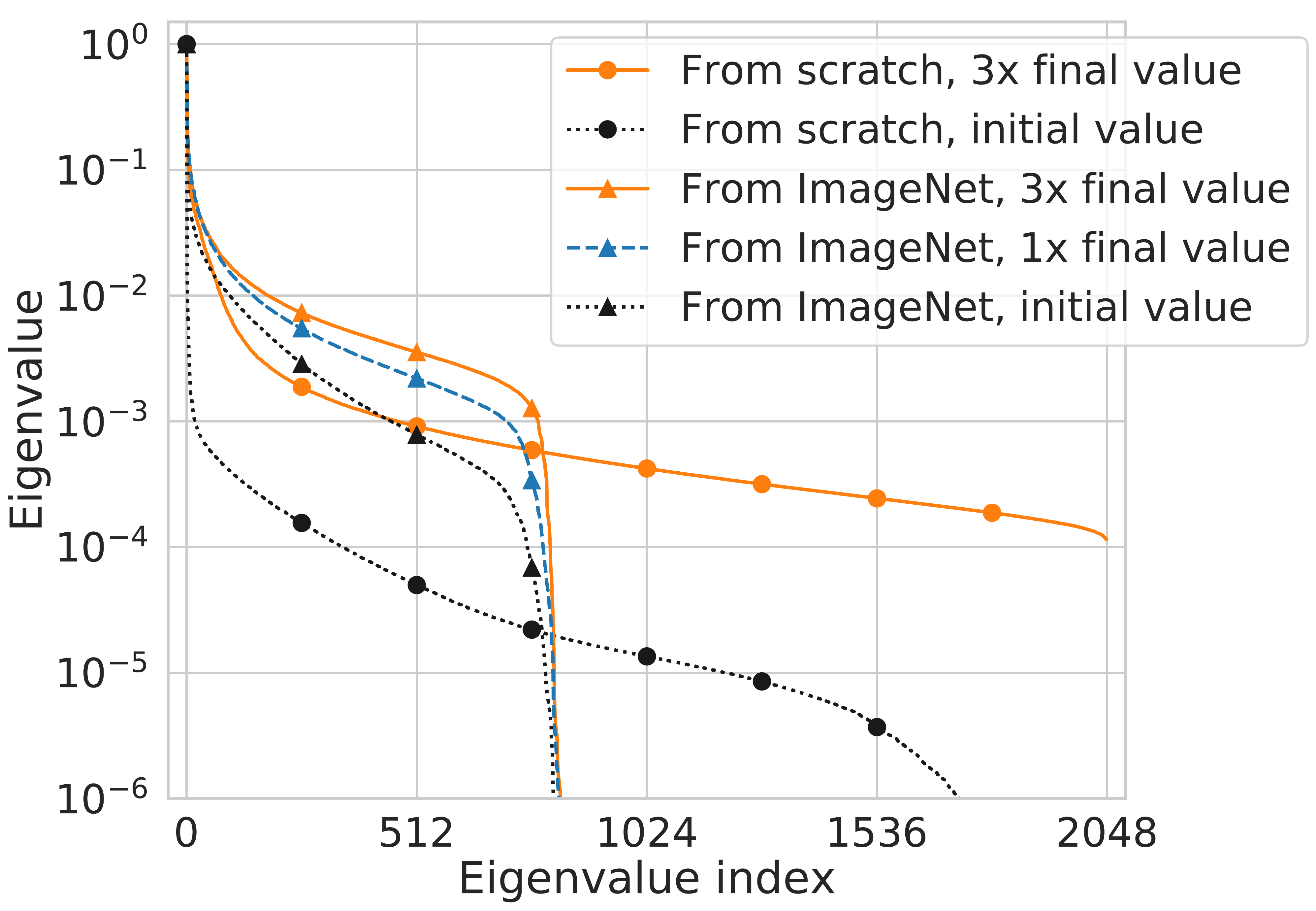}
	\end{center}
   \vspace{-3mm}
	\caption{Intrinsic dimensionalities.
		Object detectors trained from ImageNet pre-trained models \vs those trained from scratch.
		These detectors behave differently from each other even if both detectors have similar accuracy.
		\textbf{Left}: Feature maps before conv2\_1\_1.
		Lower layers in backbones converge to similar eigenspectra.
		Freezing the weights initialized from ImageNet pre-trained models in lower layers is a reasonable choice.
		\textbf{Middle}: Feature maps before conv5\_1\_1.
		Higher layers in backbones converge to dissimilar eigenspectra.
		\textbf{Right}: Feature maps before conv5\_2\_1.
		In the case of the models trained from an ImageNet pre-trained model with GN layers,
		the feature maps close to the output layer (for ImageNet classification) do not use over half of the dimensions,
		and they are not reused even after fine-tuning.
	}
	\label{fig:intrinsic_dimensionalities}
\end{figure*}

\subsection{Eigenspectrum Dynamics}
\label{sec:exp_dynamics}

To analyze the effects of pre-training for object detectors,
we observed the dynamics of the eigenspectrum of Mask R-CNN.
Figure~\ref{fig:dynamics_drop} shows the eigenspectrum of a feature map after the conv5\_1\_3
(the third convolutional layer in conv5\_1 bottleneck building block.
We call the convolutional layers of ResNet in this manner.)
of ResNet-50.
In the case of this layer, the eigenspectrum drops fast in the first 10k iterations.
Similar behavior can be seen in feature maps with 32$\times$ strides after conv5\_2\_3, conv5\_3\_3, and projection shortcut in conv5\_1.

This result demonstrates that
some information obtained in ImageNet pre-training is forgotten.
There are three possible reasons.
(i) Features for 1000-class image classification are too rich for most object detection tasks
(\eg, Classification ability needed for COCO detection is 81-class classification including a background class).
(ii) In pre-training on ImageNet, the stage 5 of ResNet is very close to the output layer.
Layers which are close to the output layer may compress information to minimum needed for the pre-training task.
(iii) The strides of conv5\_x are too coarse to localize objects.
DetNet~\cite{DetNet} and ScratchDet~\cite{ScratchDet} also discuss this problem and change the strides for object detection.
Unlike these works, our finding is that SGD (with other regularization methods) automatically limits the intrinsic dimensionalities of standard ResNet without changing the strides.

Eigenspectrum dynamics can capture not only the forgetting of pre-trained features but also the acquisition of features for COCO.
Figure~\ref{fig:dynamics_rebound} shows the numbers of eigenvalues greater than $10^{-3}$.
Eigenvalues first down, then up, in feature maps after some layers.
This rebound occurs when the learning rate decays and may relate to the learning rate schedules and a finding in~\cite{He2018Scratch} (See discussions in Sec.~\ref{sec:understanding_prior_work}).

\subsection{Intrinsic Architecture}
\label{sec:exp_intrinsic_architecture}

Here, we investigate whether
models fine-tuned from an ImageNet pre-trained model and a model trained from scratch
converge to similar intrinsic architectures.

We compare three models below.
(i) \SSS trained from scratch with $3\times$ schedule (AP$^\text{bbox}$: 39.0\%, AP$^\text{mask}$: 34.8\%),
(ii) \I trained from an ImageNet pre-trained model with $1\times$ schedule (AP$^\text{bbox}$: 38.9\%, AP$^\text{mask}$: 34.6\%), and
(iii) \III trained from an ImageNet pre-trained model with $3\times$ schedule (AP$^\text{bbox}$: 40.3\%, AP$^\text{mask}$: 35.7\%).

Figure~\ref{fig:method}(b) and Figure~\ref{fig:method}(f) show the intrinsic architectures of \I and \SSS,
and Figure~\ref{fig:intrinsic_dimensionalities} shows some characteristic intrinsic dimensionalities.
The intrinsic architecture of the model trained from scratch (\SSS) is different from that of the models trained from the ImageNet pre-trained model (\I, \III), even if the models show similar AP (\SSS \vs \I).
The accuracy of object detectors will be improved
if we properly incorporate the benefits of ImageNet pre-training and random initialization.

\begin{table*}[t]
	\setlength{\tabcolsep}{1.45mm}
	\renewcommand\arraystretch{0.95}
	\begin{center}
		\begin{tabular}{lcccccccccccc}
			\toprule
			\multirow{2}{*}{Backbone} & \multirow{2}{*}{Normalization} & \multicolumn{2}{c}{Classification} && \multicolumn{6}{c}{COCO (2$\times$ schedule)} && COCO (1$\times$) \\
			\cline{3-4}
			\cline{6-11}
			\cline{13-13}
			\addlinespace[0.4ex]
			& & MACs & \#params && AP & AP$_{50}$ & AP$_{75}$ & AP$_\textit{S}$ & AP$_\textit{M}$ & AP$_\textit{L}$ && AP \\[-0.4ex]
			\midrule
			ResNet-50~\cite{DetNet} & SyncBN & 3.8 G & --- &&
			34.5 & 55.2 & 37.7 & 20.4 & 36.7 & 44.5 && --- \\
			
			ResNet-50$^{\star}$ & GN & 4.09 G & 25.5 M &&
			35.5 & 55.6 & 38.5 & 21.3 & 37.5 & 45.3 && 29.4 \\
			
			\OurResNetSSS-50\sMACs & GN & 4.06 G & 18.6 M &&
			35.4 & 55.4 & 38.6 & 21.5 & 37.3 & 45.2 && 28.9 \\ %
			
			\OurResNetI-50\sMACs & GN & 4.05 G & 21.7 M &&
			35.5 & 55.5 & 38.6 & 21.4 & 37.3 & \textbf{46.0} && 29.2 \\
			
			\OurResNetIII-50\sMACs & GN & 4.07 G & 22.0 M &&
			35.4 & 55.6 & 38.4 & 21.3 & 37.8 & 45.5 && 29.3 \\
			
			\OurResNetIII-50\sparams\!\!$^{\star}$ & GN & 4.92 G & 24.7 M &&
			\textbf{35.8} & \textbf{55.9} & \textbf{38.9} & \textbf{21.8} & \textbf{38.0} & 45.6 && --- \\
			
			\midrule
			DetNet-59~\cite{DetNet} & SyncBN & 4.8+ G & --- && \textbf{36.3} & \textbf{56.5} & \textbf{39.3} & 22.0 & \textbf{38.4} & \textbf{46.9} && ---   \\
			DetNet-59$^{\dagger}$ & GN & 5.00+ G & 18.3+ M  && 36.2 & 56.0 & \textbf{39.3} & 22.1 & 38.3 & 46.0 && --- \\  %
			\OurDetNetII-59\sMACs & GN & 4.94+ G & 17.4+ M  && 36.2 & 56.0 & \textbf{39.3} & \textbf{22.5} & 38.1 & 46.0 && --- \\  %
			\bottomrule
		\end{tabular}
	\end{center}
	\vspace{-2mm}
	\caption{
		Efficiency on COCO object detection.
		All detectors are trained from randomly initialized weights.
		The characters like \SSS and \I in backbone names denote which models are used to determine widths.
		SyncBN: Synchronized Batch Normalization, GN: Group Normalization.
		+: Additional MACs and \#params for an additional stage out of the backbone are needed.
		$\dagger$: Our implementation.
		$\star$: We show the mean of five runs in the columns of COCO ($2\times$ schedule). The difference of COCO AP between these two backbones is statistically significant ($p<0.05$ in two-sided Welch's $t$-test).
	}
	\label{table:faster}
\end{table*}

\subsection{Discovered Backbones}

Next, we apply \OurAlgo to \SSS, \I, and \III for new backbones.
Figure~\ref{fig:method}(d) and Figure~\ref{fig:method}(h) show the architectures of \OurResNetI-50\sMACs and \OurResNetSSS-50\sMACs,
whose target resource consumption is the MACs of ResNet-50.
The architecture of \OurResNetIII-50 is similar to that of \OurResNetI-50.
Specifically, its width settings (the numbers in Figure~\ref{fig:method}(d) from below)
are (64, 64, 224, 128, 576, 256, 1152, 544, 896) for \OurResNetIII-50\sMACs,
and (64, 64, 256, 160, 608, 288, 1216, 544, 960) for \OurResNetIII-50\sparams
whose target is the number of parameters of ResNet-50.

\OurResNetI-50 and \OurResNetIII-50 have fewer widths in stage 5
and have more widths in stages 3 and 4
than ResNet.
Reducing widths in stage 5 is caused by the characteristic of models trained from ImageNet pre-trained models (Figure~\ref{fig:intrinsic_dimensionalities} Right).
Increasing widths in stages 3 and 4 may be caused by the object scales in COCO and the number of residual blocks
(The information which flows through shortcuts is stacked gradually (Figure~\ref{fig:method}(b)),
and the total amount of information may depend on the number of residual blocks).
By contrast, \OurResNetSSS-50 does not widen the widths of feature maps which pass through shortcuts in stages 3 and 4 so much.
We conjecture that this flat architecture is effective for maintaining edge information and localizing objects,
but not suitable for classification.

\subsection{Efficiency on COCO Object Detection}
\label{sec:efficiency_on_coco}

To quantify the impact on accuracy caused by the difference of intrinsic architectures
and identify better backbones than ResNet,
we trained Faster R-CNN with FPN from scratch.
Table~\ref{table:faster} shows the results.

\OurResNetSSS\sMACs, \OurResNetI\sMACs, and \OurResNetIII\sMACs, which trained with $2\times$ schedule, achieve similar AP to ResNet with fewer parameters than ResNet.
\OurResNetSSS\sMACs has $\sim$27\% fewer parameters than ResNet,
and it is the most efficient.
However, it may slightly degrade classification accuracy considering AP$_{50}$.
\OurResNetIII\sparams achieves better AP than ResNet with the similar number of parameters.
(Note that simple width multipliers~\cite{MobileNet, WRN} cannot improve AP without increasing parameters.
In addition, they degrade AP by $\sim$0.6\% to reduce parameters by $\sim$27\%.)

\OurResNetI\sMACs and \OurResNetIII\sMACs achieve higher AP than \OurResNetSSS\sMACs
if they are trained with $1\times$ schedule.
Thus, the intrinsic architectures of \I and \III have the effect of speeding up convergence.
These results are different from~\cite{He2018Scratch} because we reinitialize weights.
Besides, the differences of AP by the schedules indicate that using shorter training as a proxy task~\cite{NAS-FPN} is insufficient for this case.

In addition, we verify the effectiveness of \OurDetNetII-59\sMACs,
whose base backbone is DetNet-59 with GN (AP$^\text{bbox}$: 39.9\%) which trained from an ImageNet pre-trained model with $2\times$ schedule.
We set $10^{-3.5}$ to the threshold for eigenvalues because the number of parameters increases if it is $10^{-3}$.
Table~\ref{table:faster} shows the results.
\OurDetNetII\sMACs achieves similar AP to DetNet with $\sim$5\% fewer parameters than DetNet.
Although the parameter reduction of DetNet is more difficult than that of ResNet,
our method is also effective for DetNet.

\subsection{Efficiency on COCO Instance Segmentation}
\label{sec:efficiency_on_coco_mask}

To verify effectiveness on instance segmentation,
we also trained Mask R-CNN from scratch with $2\times$ schedule.
\OurResNetIII-50\sMACs achieves similar AP (AP$^\text{bbox}$, AP$^\text{mask}$: 36.6\%, 33.1\%)
to ResNet-50 (36.6\%, 33.0\%).
\OurResNetSSS-50\sMACs has slightly lower AP (36.5\%, 32.8\%).
This result means that the parameter reduction of Mask R-CNN is more difficult than that of Faster R-CNN,
and reflects that the intrinsic dimensionalities of networks trained on difficult tasks are large~\cite{Suzuki2018SpectralPruningCD}.

\subsection{Transferring Architecture to ImageNet}

We investigate whether \OurResNet also improves parameter efficiency
if we transfer the intrinsic architectures of the models trained on COCO to ImageNet classification.
Table~\ref{table:imagenet} shows the results.
\OurResNetSSS\sMACs has higher error rates than ResNet.
Its widths are effective for COCO but not suitable for ImageNet classification.
\OurResNetIII\sMACs achieves similar error rates to ResNet with fewer parameters than ResNet.
This result indicates that the widths of \OurResNetIII mainly depend on the redundancy inherited from an ImageNet pre-trained model (Figure~\ref{fig:intrinsic_dimensionalities} Right).

\begin{table}[t]
	\setlength{\tabcolsep}{1.2mm}
	\renewcommand\arraystretch{0.95}
	\begin{center}
		\begin{tabular}{lccc}
			\toprule
			Backbone & \#params & Top-1 err & Top-5 err \\
			\midrule

			ResNet-50$^{\star}$ & 25.5 M & 23.78\:\, & 6.97\:\, \\
			\OurResNetSSS-50\sMACs & 18.6 M & 24.71\:\, & 7.40\:\, \\ %
			\OurResNetI-50\sMACs & 21.7 M & 24.04\:\, & 7.18\:\, \\
			\OurResNetIII-50\sMACs\!\!$^{\star}$ & 22.0 M & 23.83\:\, & 6.98\:\, \\
			\OurResNetIII-50\sparams\!\!$^{\star}$ & 24.7 M & \textbf{23.45}$^*$ & \textbf{6.85}$^*$ \\

			\bottomrule
		\end{tabular}
	\end{center}
	\vspace{-2mm}
	\caption{
		Evaluation with transferring from COCO to ImageNet.
		$\star$: We show the mean of five runs. 
		*: Statistically significant differences from ResNet-50 ($p<0.05$ in two-sided Welch's $t$-test).
	}
	\label{table:imagenet}
\end{table}

\section{Discussion and Conclusions}
\label{sec:discussion}

In this section,
we first summarize our results and discuss the need to develop appropriate knowledge-transfer methods for object detectors.
After that, we discuss why architectures and learning schedules of prior work, which trains object detectors from scratch, work well.
Finally, we describe the limitations and weakness of our method.

\subsection{Appropriate Knowledge Transfer}

Although ImageNet pre-training increases intrinsic dimensionalities in higher layers (Figure~\ref{fig:method}(b)),
the increase of parameters caused by them does not improve COCO AP (Table~\ref{table:faster}).
These results do not necessarily mean that ImageNet pre-training is inefficient and meaningless for object detection.
This is because the increase of parameters in higher layers brings us better classification ability (Table~\ref{table:imagenet}).
The problem is not ImageNet pre-training itself
but rather the forgetting of ImageNet pre-trained features (Figure~\ref{fig:dynamics_drop}).
We need to take care of the compression of task-irrelevant information~\cite{michael2018on}.
Information for classification may be regarded as task-irrelevant for localization, and vice versa.

Considering the above-mentioned results, 
the current standard architectures and fine-tuning methods of object detectors are insufficient for utilizing pre-training.
For training better object detectors,
methods for appropriately transferring the knowledge of ImageNet will be needed.
The ideas of Decoupled Classification Refinement (DCR)~\cite{RevisitingRCNN_Cheng_2018_ECCV} will be helpful.
\cite{RevisitingRCNN_Cheng_2018_ECCV} decouples features for classification and localization,
and the added classifier is trained not to forget translation-invariant ImageNet pre-trained features.
To improve the efficiency of DCR,
multi-task learning with automatic branching~\cite{AdaptiveFeatureSharing_Lu_2017_CVPR} may also be needed.

\subsection{Understanding Prior Work with Our Results}
\label{sec:understanding_prior_work}

DetNet~\cite{DetNet} and ScratchDet~\cite{ScratchDet} eliminate feature maps with 32$\times$ strides from backbones,
and weigh those with finer strides relatively heavily.
These manual designs can imitate the architecture in Figure~\ref{fig:method}(h).
Considering the feature forgetting (Sec.~\ref{sec:exp_dynamics}),
the designs can avoid wasting parameters even if detectors are pre-trained.
Choosing strides automatically with~\cite{ConvolutionalNeuralFabrics, AutoDeepLab, SSL_NIPS2016, ProxylessNAS} will be more effective.

DetNet~\cite{DetNet} uses 1$\times$1 convolution projection instead of identity mapping
although stages 4, 5, and 6 have the same spatial resolution.
Our results (Figure~\ref{fig:intrinsic_dimensionalities} Right) imply that
the design keeps stages 4 and 5 away from the output layer,
and avoids too sparse representation.

Our results (Figure~\ref{fig:intrinsic_dimensionalities} Right) also imply that
current pre-training for object detectors can be considered as deep supervision~\cite{DeeplySupervisedNets}.
This is because ImageNet pre-training determines the weights of backbones only,
and the regularization effect of deep supervision remains even if the weights are fine-tuned.
Although recent work~\cite{ScratchDet, He2018Scratch} emphasizes the effectiveness of normalization layers
for training object detectors from scratch,
it is worth exploring other forms of regularization including deep supervision~\cite{DeeplySupervisedNets, DenseNet, DSOD}.

He \etal~\cite{He2018Scratch_arXiv, He2018Scratch} found that ``training longer for the first (large) learning rate is useful, but training for longer on small learning rates often leads to overfitting'' on training Mask R-CNN.
The increase of eigenspectrum in our results (Figure~\ref{fig:dynamics_rebound}) with \cite{Suzuki2018FastGE} can explain the overfitting as follows:
(i) The learning rates for training object detectors decay.
(ii) The detectors capture more detailed information about \emph{training data} by finer optimization with the small learning rates.
(iii) The eigenvalues and the intrinsic dimensionalities of the detectors increase.
(iv) The bias decreases and the variance increases.
(v) The detectors overfit if trained for longer on the small learning rates.

As described above,
eigenspectrum dynamics are useful for analyzing which feature map is responsible for what information at which time.
We believe that eigenspectrum dynamics can be a tool for analyzing neural architectures and learning rate schedules, or
early stopping by predicting generalization error with eigenspectrum of training data.

\subsection{Limitations and Weakness}

We use ResNet and its variants, FPN, and Faster/Mask R-CNN in our experiments.
It would be also interesting to conduct experiments with single-shot object detectors like SSD~\cite{SSD} and VGG-16~\cite{VGGNet} without FPN.
However, we believe that our analysis is meaningful for the computer vision community
since Faster/Mask R-CNN are standard methods for object detection and instance segmentation.

Our method can only determine the widths of feature maps.
Combining our method with compound scaling~\cite{EfficientNet} and gradient-based NAS~\cite{DARTS, ProxylessNAS, SNAS} to determine network depth, image resolution, and operations
would give us further advantages.

We only consider MACs and the number of parameters as metrics of model efficiency.
We should consider other metrics like memory footprint~\cite{MobileNetV2}, memory access cost~\cite{ShuffleNetV2},
and real latency on target platforms~\cite{NetAdapt, MnasNet, FBNet, ChamNet}.

Our method resets weights by random initialization.
This choice is practical for complicated object detectors 
because it makes codes and experiments simple.
However, applying pruning methods~\cite{PruningFilters_HaoLi_ICLR2017, YihuiHe_ChannelPruning_ICCV2017, ThiNet, Suzuki2018SpectralPruningCD} to object detectors may be better to train more efficient and accurate models.

We trained parameters after the determination of architectures in this paper.
Considering the results of recent work~\cite{NeuralRejuvenation},
the simultaneous optimization of architectures and parameters is a highly important future direction,
though the idea is classical (\eg, TWEANNs; Topology and Weight Evolving Artificial Neural Networks).
We believe that our analysis, method, and results are beneficial for the optimization
since eigenspectrum is related to both architectures and parameters.

{\small
	\bibliographystyle{ieee_fullname}
	\bibliography{shinya2019a}
}

\clearpage
\appendix
\section*{Appendix}

\section{Details of Experimental Settings}

\subsection{Experiments on COCO}

Since we use Group Normalization (GN)~\cite{GroupNormalization},
we replace box head with 4conv1fc like~\cite{GroupNormalization}.
We set the number of groups for GN to 32 (the default value in~\cite{GroupNormalization}), and 
we round the widths to multiples of 32.
(These settings are only for fair comparison.
If we change hyperparameters of GN and round to finer multiples,
we might be able to get better accuracy.)
For fair comparison with DetNet, we apply our method to backbones only.

We use Multiply-Accumulate operations (MACs) and the number of parameters as metrics of model efficiency.
The target layers for calculating the metrics are convolutional layers and fully connected layers
in the case of the backbones in our paper.

We use convolution with stride 2 in the 3$\times$3 convolutional layers of bottleneck building blocks (conv3\_1, conv4\_1, and conv5\_1) of ResNet.
Although this setting follows the setting for models with GN in Detectron~\cite{Detectron2018},
it may be the reason that the MACs and the number of parameters
of ResNet and DetNet in our implementations are slightly different from~\cite{ResNet_CVPR2016, DetNet}.

When we train models from an ImageNet pre-trained model,
we use R-50-GN.pkl\footnote{\url{https://dl.fbaipublicfiles.com/detectron/ImageNetPretrained/47261647/R-50-GN.pkl}}
provided in Detectron\footnote{\url{https://github.com/facebookresearch/Detectron/tree/master/projects/GN}}.
(R-50-GN.pkl is a ResNet-50 model trained with GN layers.
Note that the sparsity of R-50-GN.pkl is different from that of R-50.pkl\footnote{\url{https://dl.fbaipublicfiles.com/detectron/ImageNetPretrained/MSRA/R-50.pkl}}
which is a ResNet-50 model trained with batch normalization layers.
The difference relates to the \emph{dying ReLU} phenomenon and the implicit sparsity of neural networks~\cite{Yaguchi2018AdamII, Mehta2019OnIF, LeakyReLU}.)
When we train models from scratch,
weights in backbones are initialized by
He normal initialization~\cite{He2015DelvingDI} unless otherwise stated.
Weights in FPN are initialized by
Glorot uniform initialization~\cite{Glorot2010UnderstandingTD}.
These settings follow Detectron~\cite{Detectron2018}.

The weights of DetNet and \OurDetNet are initialized by the default initialization method of PyTorch 0.4.0\footnote{\url{https://github.com/pytorch/pytorch/blob/v0.4.0/torch/nn/modules/conv.py\#L40-L47}}.
This is because DetNet-59 whose weights are initialized by He normal initialization has $0.5\%$ lower COCO AP in our experiment.
For simplicity and fair comparison with ResNet, the $P_6$ of FPN is not used by Fast R-CNN heads like~\cite{FPN}.

We verified that Detectron.pytorch~\cite{Detectron_pytorch} can reproduce the results of Detectron before conducting experiments in our paper.
Strictly speaking, we verified the reproducibility with Faster R-CNN, Mask R-CNN and Keypoint R-CNN (Mask R-CNN for human pose estimation~\cite{Mask_R-CNN}) with FPN and ImageNet pre-trained ResNet-50 whose batch normalization layers are frozen.

``AP'', which is the primary metric of COCO, means AP$_\text{IoU=.50:.05:.95}$
(Average Precision averaged over the multiple Intersection-over-Union),
and AP without superscript means AP$^\text{bbox}$ (AP for object detection) in our paper.

When we calculate eigenspectra, we first randomly sample 5,000 images from COCO \texttt{train2017} set for fast calculation, then feed-forward the sampled images.
The images are resized such that their shorter side is 800 pixels~\cite{FPN}.
The resizing is the same for training and testing.

See the codes of
Detectron.pytorch\footnote{\url{https://github.com/roytseng-tw/Detectron.pytorch/commit/8315af319cd29b8884a7c0382c4700a96bf35bbc}}
for other implementation details.

\subsection{Experiments on ImageNet}

When we transfer the intrinsic architectures of the models trained on COCO to ImageNet classification,
we train models with batch normalization layers for 100 epochs.
The initial learning rate is 0.1,
and the learning rate is reduced by $10\times$
at 30, 60, and 90 epochs.
We use synchronous SGD
with an effective batch size of 256 (= 64 images/GPU $\times$ 4 GPUs),
a momentum of 0.9, and a weight decay of $10^{-4}$.
We crop input images to 224$\times$224 pixels.

See the codes of Neural Network Distiller\footnote{\url{https://github.com/NervanaSystems/distiller/commit/a89b3ad19da164f517e5b9e9e568c94069cc0c83}}
for other implementation details
(Although Distiller is a library for neural network compression, we use Distiller only for calculating MACs and the number of parameters).

\end{document}